\pdfoutput=1
\documentclass[11pt]{article}

\usepackage[final]{acl}

\usepackage{times}
\usepackage{latexsym}

\usepackage[T1]{fontenc}
\usepackage[utf8]{inputenc}

\usepackage{microtype}

\usepackage{inconsolata}

\usepackage{graphicx}
\usepackage{hyperref}
\usepackage{booktabs} 
\usepackage{url}
\usepackage{pdflscape}
\usepackage{subfigure}
\usepackage{tikz}
\usepackage{graphicx}
\usepackage{makecell}
\usepackage{algorithm}
\usepackage{algorithmic}
\usepackage{makecell}
\usepackage{amsmath}
\usepackage{wrapfig}
\usepackage{pifont} % 符号支持
\usepackage{svg}
\usepackage{listings}
\usepackage{xcolor}  % 可以用来定义代码的颜色
\usepackage[skins, breakable]{tcolorbox}
\newtcolorbox{cvbox}[1][]{
    enhanced,
    after skip=8mm,%   enlarge distance to the next box
    title=#1,
    breakable = true,
    fonttitle=\sffamily\bfseries\color{white},
    coltitle=white,
    colbacktitle=black!75!white, %gray!10,   % <- defines background color in title
    titlerule= 0pt,         % <- sets rule underneath title 
    overlay={%
        \ifcase\tcbsegmentstate
        \or%
        \else%
        \fi%
    }
    colback = gray!5!white,         % <- defines background color in box
    colframe = black!75     % <- defines color of frame
    }
    
\usepackage{amsmath,amsfonts,bm}

\def\eqref#1{equation~\ref{#1}}
\def\1{\bm{1}}

\def\vk{{\bm{k}}}

\def\vq{{\bm{q}}}

\def\vv{{\bm{v}}}

\def\vx{{\bm{x}}}

\def\mA{{\bm{A}}}

\def\mK{{\bm{K}}}

\def\mQ{{\bm{Q}}}

\def\mV{{\bm{V}}}
\def\mW{{\bm{W}}}
\def\mX{{\bm{X}}}

\DeclareMathAlphabet{\mathsfit}{\encodingdefault}{\sfdefault}{m}{sl}
\SetMathAlphabet{\mathsfit}{bold}{\encodingdefault}{\sfdefault}{bx}{n}

\def\gA{{\mathcal{A}}}

\def\gC{{\mathcal{C}}}

\def\gI{{\mathcal{I}}}

\def\gM{{\mathcal{M}}}

\def\gS{{\mathcal{S}}}

\def\gW{{\mathcal{W}}}
\def\gX{{\mathcal{X}}}
\def\gY{{\mathcal{Y}}}

\newcommand{\redcross}{{\color{red}\ding{55}}}
\newcommand{\greentick}{{\color{green}\ding{51}}}

\title{Boosting Long-Context Management via Query-Guided Activation Refilling }

\author{Hongjin Qian$^{1}$, Zheng Liu$^1$\thanks{Corresponding author.}, Peitian Zhang$^2$, Zhicheng Dou$^2$, Defu Lian$^3$\\
        $^1$ Beijing Academy of Artificial Intelligence \\ 
        $^2$  Gaoling School of Artificial Intelligence, Renmin University of China\\
        $^3$ University of Science and Technology of China \\
        \texttt{\{chienqhj,zhengliu1026\}@gmail.com} \\
}

\begin{document}
\maketitle
\begin{abstract}
Processing long contexts poses a significant challenge for large language models (LLMs) due to their inherent context-window limitations and the computational burden of extensive key-value (KV) activations, which severely impact efficiency. For information-seeking tasks, full context perception is often unnecessary, as a query's information needs can dynamically range from localized details to a global perspective, depending on its complexity. However, existing methods struggle to adapt effectively to these dynamic information needs.

In the paper, we propose a method for processing long-context information-seeking tasks via query-guided \textbf{AC}tivation \textbf{RE}filling (ACRE). ACRE constructs a \textit{Bi-layer KV Cache} for long contexts, where the layer-1 (L1) cache compactly captures global information, and the layer-2 (L2) cache provides detailed and localized information. ACRE establishes a proxying relationship between the two caches, allowing the input query to attend to the L1 cache and dynamically refill it with relevant entries from the L2 cache. This mechanism integrates global understanding with query-specific local details, thus improving answer decoding.
Experiments on a variety of long-context information-seeking datasets demonstrate ACRE's effectiveness, achieving improvements in both performance and efficiency. We will release our source codes in \href{https://github.com/qhjqhj00/activation_refilling}{\textit{this repository}}.
\end{abstract}

\section{Introduction}
Recently, large language models (LLMs) have become widely used for daily information-seeking tasks, such as ChatGPT~\cite{gpt-4}. However, their capabilities are inherently limited by the difficulty of updating parametric knowledge. To address this, incorporating external knowledge as a context has become a common approach~\cite{zhao2024surveylargelanguagemodels}. In practice, this external knowledge often involves long contexts, such as long documents or novels, which pose significant challenges due to the large KV activations accumulated during inference, demanding substantial computational resources and reducing efficiency~\cite{longctx, bai2023longbench, zhang2024inftybench}.

\begin{figure}
    \centering
    \includegraphics[width=\linewidth]{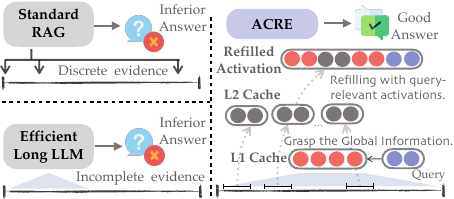}
    \caption{Comparison of ACRE, standard RAG, and efficient long LLMs for information-seeking tasks. Standard RAG retrieves evidence without full-context perception, and long LLMs struggle with contexts exceeding their native window. ACRE overcomes these limitations with a resource-efficient bi-layer KV cache and query-guided refilling, capturing both global and local information while enhancing performance.}
    \label{fig:case}
\end{figure}

To address the challenges posed by excessive KV activations, previous works have proposed various strategies: reducing the precision of activation tensors~\cite{liukivi,xu2024think}, dividing long contexts into smaller chunks for independent processing~\cite{lee2024humaninspiredreadingagentgist, yoon2024compactcompressingretrieveddocuments}, or compressing KV activations into shorter representations through selection or sparse attention~\cite{zhang2023h2oheavyhitteroracleefficient, li2024snapkv, xiao2024efficientstreaminglanguagemodels, jiang2024minference}.
Retrieval-Augmented Generation~(RAG) has also emerged as a promising approach, retrieving precise evidence from long contexts to support answer generation~\cite{gao2024retrievalaugmented}. 

However, most existing methods follow a unilateral strategy: either compromising the semantic richness of KV activations to create compact global representations, such as with quantized activations~\cite{liukivi}, or concentrating solely on detailed local information, such as RAG methods~\cite{gao2024retrievalaugmented}. Moreover, most lightweight KV methods remain constrained by the native context length limit, leading to significant performance degradation when processing contexts that exceed this limit~\cite{zhang2024extending,}.

In information-seeking tasks, we argue that the information needs of a user query can dynamically range from localized details to a global perspective, depending on the query’s complexity. For instance, given a novel, the query “What are the main characters’ names?” involves localized information needs and can be answered using specific local evidence. In contrast, the query “How do the main characters drive the story’s development?” requires a global understanding of the entire book~\cite{qian2025hawkbench}.

To address dynamic information needs in information-seeking tasks, we propose ACRE, a method that employs a bilateral strategy to capture a global perspective across the full context and enhance local details using query-guided activation refilling. Figure~\ref{fig:case} presents an overview of ACRE’s framework along with a comparison against efficient long LLMs and RAG methods.

Specifically, ACRE constructs a bi-layer KV activation cache for long contexts, comprising an L1 cache and an L2 cache. The L1 cache captures compact yet global information from the full context, while the L2 cache retains localized, detailed information. Notably, the L1 cache is significantly smaller than the L2 cache. During the forward pass of the LLM, the L1 and L2 caches are interleaved into a nested structure, with each L1 tensor optimized to proxy the semantics of its corresponding L2 cache.
To enhance efficiency, we replace the original full attention mechanism—where each token attends to all preceding tokens—with a tailored selective attention mechanism. In this approach, tokens perform full attention on recent L1 and L2 tokens but only attend to distant L1 tokens. This selective attention mechanism significantly reduces computational costs, enabling ACRE to process long contexts more efficiently.

After the forward pass, the nested KV cache is decomposed back into separate L1 and L2 caches. For an input query, ACRE first uses the query to attend to the compact L1 cache. Based on the resulting attention score distribution, ACRE selectively refills key entries of the L1 cache with the corresponding L2 cache entries, thereby enriching local details. This process is referred to as query-guided activation refilling.

ACRE is trained through an efficient two-stage process. The first stage focuses on constructing the bi-layer KV cache, while the second stage targets query-guided activation refilling. Throughout both stages, ACRE updates only a small subset of model parameters, ensuring training efficiency.

We evaluate ACRE across a wide range of long-context information-seeking tasks~\cite{bai2023longbench,zhang2024inftybench,qian2024memoragmovingnextgenrag}. The experimental results confirm the effectiveness of ACRE. Our key contributions are summarized as follows:
(1) We design a flexible and efficient bi-layer KV activation cache mechanism for long contexts, which captures compact global information while preserving local details.
(2) We introduce ACRE, a method that leverages the bi-layer KV activation cache with a query-guided activation refilling mechanism to efficiently handle long-context information-seeking tasks.
(3) We demonstrate that ACRE achieves superior performance on long-context information-seeking tasks, effectively handling contexts much longer than LLMs’ typical context limits, while substantially reducing computational resources and latency.

\section{Method}
\label{sec:method}

\begin{figure*}
    \centering
    \includegraphics[width=\linewidth]{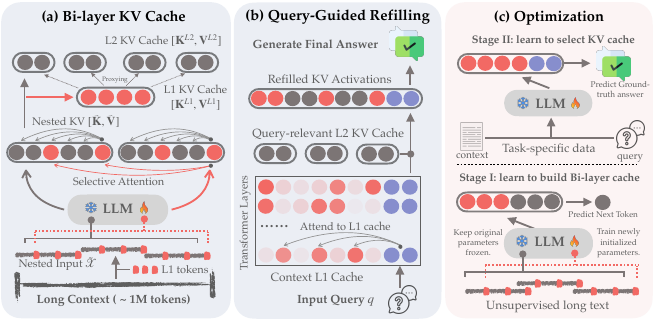}
    \caption{Overview of ACRE. (a) ACRE constructs the Bi-layer KV cache from a long context. (b) For an input query, ACRE refills the L1 KV cache with query-relevant entries from the L2 KV cache and decodes the final answer based on the refilled cache. (c) The two-stage optimization process used to train ACRE is illustrated.}
    \label{fig:method}
\end{figure*}
\subsection{Preliminary}
The process of solving information-seeking tasks using LLMs can be succinctly described as $\gY = \gM(\gX)$,  where $\gM(\cdot)$ denotes the LLM, $\gY$ represents the output answer and $\gX$ represents the input sequence. $\gX$ can take various forms, ranging from a standalone query to a complex instruction prompt. In this paper, we focus on information-seeking tasks with long contexts. Therefore, we define the input sequence $\gX$ as comprising a query $q$ and a long context $\gC$, denoted by $\gX = (\gC, q)$. 
    
For the input $\gX$, a Transformer-based LLM computes multi-head attention (MHA) as follows:
\begin{align}
    \mQ &= \mX \cdot \mW_{Q}, \\
    \mK &= \mX \cdot \mW_{K}, \\
    \mV &= \mX \cdot \mW_{V}, \\
    \gA(\mQ, \mK, \mV) &= \text{softmax}\left(\frac{\mQ \cdot \mK^\top}{\sqrt{d}}\right) \cdot \mV,
    \label{eq:attention}
\end{align}
where $\mX$ represents the hidden states of the input sequence $\gX$, and $\mW_{Q}$, $\mW_{K}$, and $\mW_{V}$ are the projection weight matrices for the query $\mQ$, key $\mK$, and value $\mV$, respectively~\cite{vaswani2023attentionneed}. The attention function $\gA(\cdot)$ is applied iteratively across multiple layers and attention heads. For simplicity, we omit the layer and head indices.
 % TODO
 
The inference process of LLMs can be divided into two stages: (1)~prefilling and (2)~decoding~\cite{liukivi}. During the prefilling stage, the input sequence $\gX$ is processed through each layer using MHA, and the layer-wise key-value activations $[\mK, \mV]$ are cached. These cached activations are reused in the decoding stage to avoid redundant computations, enabling efficient processing. However, as MHA computation has quadratic complexity with respect to the sequence length \( n \), handling long contexts becomes computationally expensive. This often results in slow processing speeds and out-of-memory issues, particularly when dealing with long input contexts~\cite{dong2023survey}. 

To address the challenges posed by oversized KV caches for long contexts, we propose ACRE, a framework that constructs a \textit{Bi-layer KV Cache} and employs a \textit{Query-Guided Refilling} mechanism to enable a flexible KV cache that captures both global context and query-specific local details, ensuring efficient and high-quality answer decoding.

\subsection{Overview of ACRE}
Figure~\ref{fig:method} provides an overview of ACRE. Specifically, for a information-seeking task with a long context \( \gC \), ACRE organizes the long context into a bi-layer KV activation cache during the pre-filling stage, as shown in Figure~\ref{fig:method}~(a).

The construction of the \textit{Bi-layer KV Cache} begins by interleaving newly introduced L1 tokens into the input context. Through model forwarding, a nested KV cache $[\tilde{\mK}, \tilde{\mV}]$ is obtained. This nested KV cache is then decomposed into a Bi-layer KV cache: the layer-1 (L1) cache, which is compact and stores global information from the full long context, and the layer-2 (L2) cache, which holds detailed and localized information. Each tensor in the L1 cache serves as a semantic proxy for a corresponding sequence of tensors in the L2 cache.

We denote the L1 KV cache as $[\mK^{L1}, \mV^{L1}] \in \mathbb{R}^{m \times d}$ and the L2 KV cache as $[\mK^{L2}, \mV^{L2}] \in \mathbb{R}^{n \times d}$. Here, the length of the L1 KV cache, \( m \), is significantly smaller than \( n \), the length of the L2 KV cache. To optimize memory usage, the L2 cache can be offloaded to CPU memory, while the L1 cache is retained in GPU memory as a constant cache after constructing the bi-layer KV cache. This design significantly improves memory efficiency in practical applications.

The \textit{Bi-layer KV Cache} is constructed exclusively for input contexts, enabling it to be reused across different information-seeking tasks that share the same context. Given an input query \( q \), ACRE utilizes \( q \) to attend to the L1 cache, computing attention scores. Based on these scores, ACRE selectively refills the L1 cache by retrieving the most informative entries from the L2 cache, which are proxied by the corresponding most attentive L1 cache tensors. This process recovers a partial nested cache to support answer decoding and is referred to as \textit{query-guided activation refilling}, which is shown in Figure~\ref{fig:method}~(b). 

By leveraging both the L1 KV cache and the query-specific L2 KV cache, the final KV cache captures global information from the full long context while preserving local details. This design significantly enhances the performance of long-context information-seeking tasks. In the following sections, we provide the technical details of ACRE.

\subsection{Bi-Layer KV Cache}
To construct the bi-layer KV cache, we introduce a new type of token, called L1 tokens, denoted as $\gX^{L1} = (x^{L1}_1, \cdots, x^{L1}_m)$. The original tokens of the input sequence are referred to as L2 tokens, denoted as $\gX^{L2} = (x_1, \cdots, x_n)$. 
By interleaving the L1 and L2 tokens, the input sequence $\gX$ is transformed into a nested sequence $\tilde{\gX}$:
\begin{align}
    \tilde{\gX} = (x_1, \cdots, x_l, x^{L1}_1, x_{l+1}, \cdots, x_n, x^{L1}_m),
\end{align}
where each L1 token is inserted after every $l$ L2 tokens, acting as a semantic proxy for the preceding $l$ L2 tokens. We refer to $l$ as the L1/L2 interval. For the L1 tokens, we initialize an additional set of trainable weight matrices $\mW_Q^{L1}$, $\mW_K^{L1}$, and $\mW_V^{L1}$, while keeping the original weight matrices for L2 tokens frozen.

After constructing the nested sequence $\tilde{\gX}$, we adapt the attention computation defined in Eq.~(\ref{eq:attention}). Specifically, for the key $\mK$, the original projection $\mK = \mX \cdot \mW_K$ is replaced with:
\begin{align}
    \mK = 
    \begin{cases} 
        \vx \cdot \mW_{K}^{L1}, & \text{if } x \text{ is an L1 token}, \\
        \vx \cdot \mW_{K}, & \text{if } x \text{ is an L2 token},
    \end{cases} 
\end{align}
where $\vx \in \mX$. Through multi-head attention, this modification yields the nested key activations:
\begin{align}
    \tilde{\mK} = [\vk_1, \cdots, \vk_l, \vk^{L1}_1, \cdots, \vk_n, \vk_m^{L1}].
    \label{eq:nested_kv}
\end{align}
Similarly, the nested value activations $\tilde{\mV}$ are computed as:
\begin{align}
    \tilde{\mV} = [\vv_1, \cdots, \vv_l, \vv^{L1}_1, \cdots, \vv_n, \vv_m^{L1}].
\end{align}

By decomposing the nested KV cache, we obtain the bi-layer KV cache as follows:
\begin{gather}
    \mK^{L1} = [\vk_1^{L1}, \cdots, \vk_m^{L1}], \\
    \mV^{L1} = [\vv_1^{L1}, \cdots, \vv_m^{L1}], \\
    \mK^{L2} = \big[ \underbrace{\vk_1, \cdots, \vk_l}_{\vk_1^{L1}}, \cdots, \underbrace{\vk_{n-l}, \cdots, \vk_n}_{\vk_m^{L1}} \big], \\
    \mV^{L2} = \big[ \underbrace{\vv_1, \cdots, \vv_l}_{\vv_1^{L1}}, \cdots, \underbrace{\vv_{n-l}, \cdots, \vv_n}_{\vv_m^{L1}} \big],
\end{gather}
where $\underbrace{\vk_1, \cdots, \vk_l}_{\vk_1^{L1}}$ represents the proxying relationship between the L1 cache and the L2 cache.

As previously mentioned, directly computing full attention over the long sequence $\gX$ is both computationally expensive and resource-intensive. To efficiently construct the bi-layer KV cache, we propose a \textit{selective attention mechanism}. This mechanism maintains a relatively small working context window $\gW$, enabling current tokens to perform full attention on recent L1 and L2 tokens while only attending to distant L1 tokens. For instance, when computing KV activations at step~$n$, we prune the previous KV cache $[\tilde{\mK}, \tilde{\mV}]$ as follows:
\begin{gather}
    \tilde{\mK} = [\vk^{L1}_1, \cdots, \vk^{L1}_i, \vk_j, \cdots, \vk_n,\vk_m^{L1}], \\
    \tilde{\mV} = [\underbrace{\vv^{L1}_1, \cdots, \vv^{L1}_i}_{\text{distant L1 tokens}}, \underbrace{\vv_j, \cdots, \vv_n, \vv_m^{L1}}_{\text{recent L1 / L2 tokens}}],
\end{gather}
subject to the constraints $\mid\tilde{\mK}\mid \leq \gW$ and $\mid\tilde{\mV}\mid \leq \gW$. 
Through this mechanism, we sequentially process the full sequence $\tilde{\gX}$ into KV activations using a short working context window, achieving both high computational efficiency and economical memory usage.

\subsection{Query-Guided Activation Refilling}
After constructing the bi-layer KV cache for the context, we obtain the L1 KV cache $[\mK^{L1}, \mV^{L1}]$, which serves as a global yet compact representation of the full long context, and the L2 KV cache $[\mK^{L2}, \mV^{L2}]$, which provides detailed but memory-intensive representations. To optimize memory usage, the L1 KV cache is retained as a constant cache in GPU memory, while the L2 KV cache is offloaded to CPU memory.

For an input query $q$, relying solely on the L1 KV cache is feasible but lacks query-specific detailed information. To address this limitation, ACRE proposes refilling the compact L1 KV cache with selected entries from the L2 KV cache that are most relevant for answering the query.
Specifically, the query state $\mQ_q$ for the input query $q$ is computed as $\mQ_q = \vq \cdot \mW_Q$. Using this query state, the attention distribution is calculated as:
$
    \mA = \text{softmax}\left(\frac{\mQ_q \cdot {\mK^{L1}}^\top}{\sqrt{d}}\right),
$
where $\mA \in \mathbb{R}^{h \times m \times t}$, $h$ is the number of attention heads, $m$ is the length of L1 cache, and $t$ is the length of the query $q$. The attention scores $\gS$ are then obtained by applying mean pooling:
\begin{align}
    \gS = \text{Pool}_{\text{dim}=0,2}(\mA), \quad \gS \in \mathbb{R}^m,
\end{align}
where $\gS$ serves as a guiding signal to select relevant entries from the L2 KV cache. The selection process is defined as:
\begin{gather}
    \gI = \operatorname{arg\,top}_k(\gS), \\ 
    k = \left\lfloor \frac{\min(\gW - m, \eta)}{l} \right\rfloor,
\end{gather}
where $k$ is dynamically determined based on the maximum length of the predefined working context window $\gW$ or the maximum refilling length $\eta$, and $\gI$ represents the set of selected indices.

After selection, the L1 KV cache is refilled with the chosen entries from the L2 KV cache. For example, if $\gI = \{2\}$, the refilled KV cache becomes:
\begin{align}
    \mK &= [\vk^{L1}_1, \vk_{l+1}, \cdots, \vk_{2l}, \vk^{L1}_2, \cdots, \vk_m^{L1}], \\
    \mV &= [\vv^{L1}_1, \underbrace{\vv_{l+1}, \cdots, \vv_{2l}}_{\text{refilled L2 KV cache}}, \vv^{L1}_2, \cdots, \vv_m^{L1}].
\end{align}
This refilling process is performed independently for each layer. With the refilled KV cache, ACRE decodes the final answer $\gY$ in a standard auto-regressive manner.

\subsection{Model Optimization}
ACRE is characterized by its \textit{Bi-layer KV Cache} structure and \textit{Query-Guided Activation Refilling} mechanism. Its effectiveness relies on two key abilities: (1) the L1 KV activations must faithfully represent the L2 KV activations, and (2) given an input query $q$, the most relevant L2 KV activations must be efficiently retrieved. To optimize these abilities, we employ a two-stage optimization strategy.

In stage 1, the objective is to maximize the semantic volume of the L1 KV activations to effectively represent the corresponding L2 KV activations. This is achieved by predicting the next token using the previously accumulated L1 tokens and the recent L2 tokens. The optimization can be expressed through a cross-entropy loss:
\begin{align}
    \mathcal{L}_{\text{stage-1}} = - \sum_{t=1}^{T} \log \mathcal{P}(x_t \mid x^{L1}_{[1:i]}, x_{[j:t-1]}),
\end{align}
where $x^{L1}_{[1:i]}$ denotes the accumulated L1 tokens, and $x_{[j:t-1]}$ denotes the recent L2 tokens.

In stage 2, the objective is to enable ACRE to retrieve the most relevant L2 KV activations for refilling the L1 KV cache based on an input query $q$. Since the L2 KV cache is proxied by the L1 KV cache, accurately attending to the most useful L1 KV activations allows retrieval of the corresponding L2 KV activations via the proxying relationship. 
To achieve this, we optimize ACRE using task-specific data comprising long contexts and input queries. The optimization employs the following loss function:
\begin{align}
    \mathcal{L}_{\text{stage-2}} = - \sum_{t=1}^{T} \log \mathcal{P}(y_t \mid \gX^{L2}, q),
\end{align}
where $y$ represents the ground-truth answer, and $q$ is the input query. This loss ensures that ACRE learns to produce accurate answers solely based on the L1 KV cache while maintaining its ability to retrieve the most relevant L2 KV activations.

\section{Experiments}
\subsection{Dataset}
We evaluate ACRE and all baseline models across 12 information-seeking tasks from three public long-context benchmarks: LongBench~\cite{bai2023longbench}, InfiniteBench~\cite{zhang2024inftybench}, and UltraDomain~\cite{qian2024memoragmovingnextgenrag}. These 12 datasets are categorized as follows:
(1)~\textbf{Complex QA}~\cite{qian2024memoragmovingnextgenrag}: Financial, Legal, Physics, Biology, Math, and CS. These tasks involve practical, high-level queries with extra-long contexts spanning specialized domains. Many queries demand a global and in-depth understanding of the full context, making them especially challenging.
(2)~\textbf{Single-Document QA}: NarrativeQA~\citep{kočiský2017narrativeqa}, Qasper~\citep{dasigi2021dataset}, MultiFieldQA~\citep{bai2023longbench}, and En.QA~\cite{zhang2024inftybench}.
(3)~\textbf{Multi-Document QA}:  2WikiMQA~\citep{ho-etal-2020-constructing}, and MuSiQue~\citep{trivedi2022musique}.

\begin{table*}[t]
    \centering
    \small
    \caption{Main experimental results. The best results are in bold, and the second-best are \underline{underlined}. All methods use Qwen2.5-3B-Instruct as the underlying LLM. Baselines in the second block directly process the full context, while those in the third block divide the context into chunks and find evidence using a retriever. In the second row, $\text{ave}(|\gC|) ($k$)$ means the average context length. }
    
\begin{tabular}{lcccccccc|cccc}
\toprule
Dataset  & nar & fin & legal  & phy & bio & en.qa  & math &   cs &  qas&  mul & 2wiki&mus \\
$\text{ave}(|\gC|) ($k$)$& 18.4 &40.6 &51.4 & 105.8 & 125.3 & 192.6 & 197.9  &215.9  & 3.6 & 4.6 & 4.9  &  11.2\\

\midrule
 & \multicolumn{8}{c}{\textsc{Ave. Context Length > 16k}} & \multicolumn{4}{c}{\textsc{Ave. Length < 16k}} \\
\midrule
Original & 22.0& 36.8 & 42.6 & \underline{38.2} & \underline{35.8} & \underline{20.1} & \underline{36.3} & \underline{35.6} & 37.4 & 48.5 & 36.3  & 22.1\\
\midrule
KIVI &21.1 & 27.0 &39.5 &  35.3 & 33.2 & 15.6 & 32.1 & 33.4  & 37.1 & 46.1 & 35.0  & 22.1\\
Beacon &20.2 & 37.8 & 43.9 & 37.1 & 33.7 & 18.3 & 31.8 & 32.3 & 30.4 & 35.6 & 24.7& 24.7\\
SelfExtend & 20.8  & 37.5 & 40.0 & 29.1 & 29.9 & 11.4 & 31.6 & 30.4 & 36.0 & \underline{49.6} & 37.1  & 25.1\\
StreamingLLM  & 18.8 & 27.3 & 26.2 & 31.4 & 27.4 & 8.3 &  30.0 & 26.9 & 33.4 & 38.6 & 32.1  & 12.2\\
MInference& 22.2& 35.6 & 37.2 & 32.9 & 28.5 & 8.9 & 30.3 & 27.1 &36.2 &  48.6& 36.0  & 23.5\\

\midrule
RAG &18.9 & 36.9 & 38.6 & 22.1 & 18.4 & 11.3 & 19.2 & 19.3& \underline{38.6}& 46.6 & \underline{37.8}  & 20.8\\
RQRAG &19.0 & 37.0 & 39.0 & 28.0&  23.0  & 12.0  & 26.1 & 24.1 &37.6& 47.3 & 37.4  & 21.8\\
MemoRAG &\underline{24.0} & \underline{41.5} & \underline{44.8} & 36.9 &  33.2 & 13.2 & 33.1 & 33.4 & 34.1 & 49.1 & \textbf{38.0}  & \underline{26.0} \\
\midrule

ACRE & \textbf{27.8}  & \textbf{46.4} & \textbf{47.7} & \textbf{41.6} & \textbf{38.3} & \textbf{23.6} & \textbf{41.9} & \textbf{45.9} &\textbf{39.6} & \textbf{50.0} & 36.4  & \textbf{26.2}\\

\bottomrule
\end{tabular}
\label{tab:qa}
\end{table*}

\subsection{Baseline Models}
We compare ACRE with the following baselines: 
\textbf{Original}: Directly fits the maximum context length of the underlying LLMs.  
\textbf{KIVI}~\cite{liukivi}: Quantizes KV activations into 4-bit precision.  
\textbf{Beacon}~\cite{zhang2024soaring}: Compresses the full KV activations into beacon activations.  
\textbf{SelfExtend}~\cite{jin2024llmmaybelonglmselfextend}: Applies hierarchical positional encoding to extend the model's context window.  
\textbf{MInference}~\cite{jiang2024minference}: Dynamically applies different sparse attention mechanisms across all attention heads.  
\textbf{StreamingLLM}~\cite{xiao2024efficientstreaminglanguagemodels}: Attends only to recent tokens and sink tokens.  
\textbf{RAG}: Uses standard RAG pipelines to retrieve relevant evidence from the full context.  
\textbf{RQRAG}~\cite{chan2024rqraglearningrefinequeries}: Rewrites the input query into sub-queries and retrieves evidence for each sub-query.   
\textbf{MemoRAG}~\cite{qian2024memoragmovingnextgenrag}: Applies a memory model to form a compact global memory over the full context, providing answer clues that assist the retrieval process for better evidence retrieval.  

In the main experiments (Section~\ref{sec:main}), we use Qwen2.5-3B-Instruct as the underlying model. To analyze the impact of using different underlying models,  we also experiment with Llama3.2-3B-Instruct and Qwen2.5-7B-Instruct in Section~\ref{sec:abl}. All three LLMs have a native context window of 128K~\cite{qwen2, grattafiori2024llama3herdmodels}. The implementation details of ACRE and all baselines are in Appendix~\ref{sec:imp}.

\subsection{Main Results}
\label{sec:main}
In Table~\ref{sec:main}, we present the results of the main experiments, demonstrating that ACRE outperforms all baselines across most datasets. These results highlight the effectiveness of ACRE’s design. Specifically, we derive the following findings: 
(1)~ACRE consistently outperforms the baseline approach of feeding the full context directly into LLMs. This improvement stems not only from ACRE’s ability to process contexts exceeding the native LLM’s context window but also from its precise focus on query-relevant local information, effectively filtering out irrelevant details through query-guided activation refilling.
(2)~Baselines in the second block generally perform worse than directly feeding the full context into LLMs. This is attributed to semantic loss caused by compressing full KV activations. In contrast, ACRE leverages its bi-layer KV cache and query-guided activation refilling to recover local detailed semantics from the L2 cache that are absent in the L1 cache, resulting in superior performance.
(3)~Baselines in the third block use retrieval tools to extract precise evidence from long contexts. While effective for queries with clear information needs, these methods struggle with complex queries that require a higher-level understanding of the full context. ACRE overcomes this limitation by utilizing the global information in the L1 cache and dynamically refilling it with query-relevant local details from the L2 cache, thereby adapting to the varying information needs of different queries.

\subsection{Ablation Study}
\label{sec:abl}
To thoroughly validate the effectiveness of our method design, we perform detailed ablation studies as follows:

\begin{figure}
    \centering
    \includegraphics[width=0.95\linewidth]{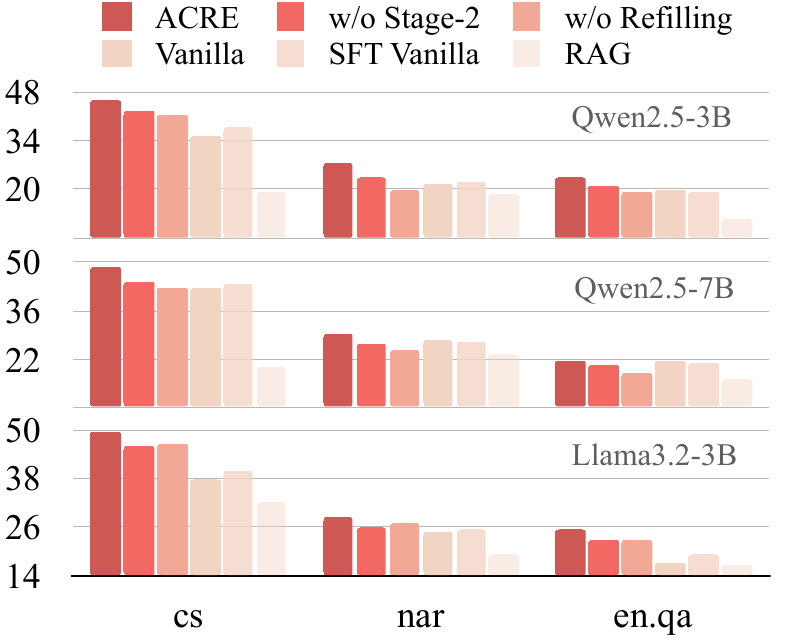}
    \caption{Ablation Study on Model Design Variations Across Different LLMs.}
    \label{fig:abl}

\end{figure}

\begin{figure}[t]
    \centering
    
    % \hfill
    \subfigure{
        \includegraphics[width=0.225\textwidth]{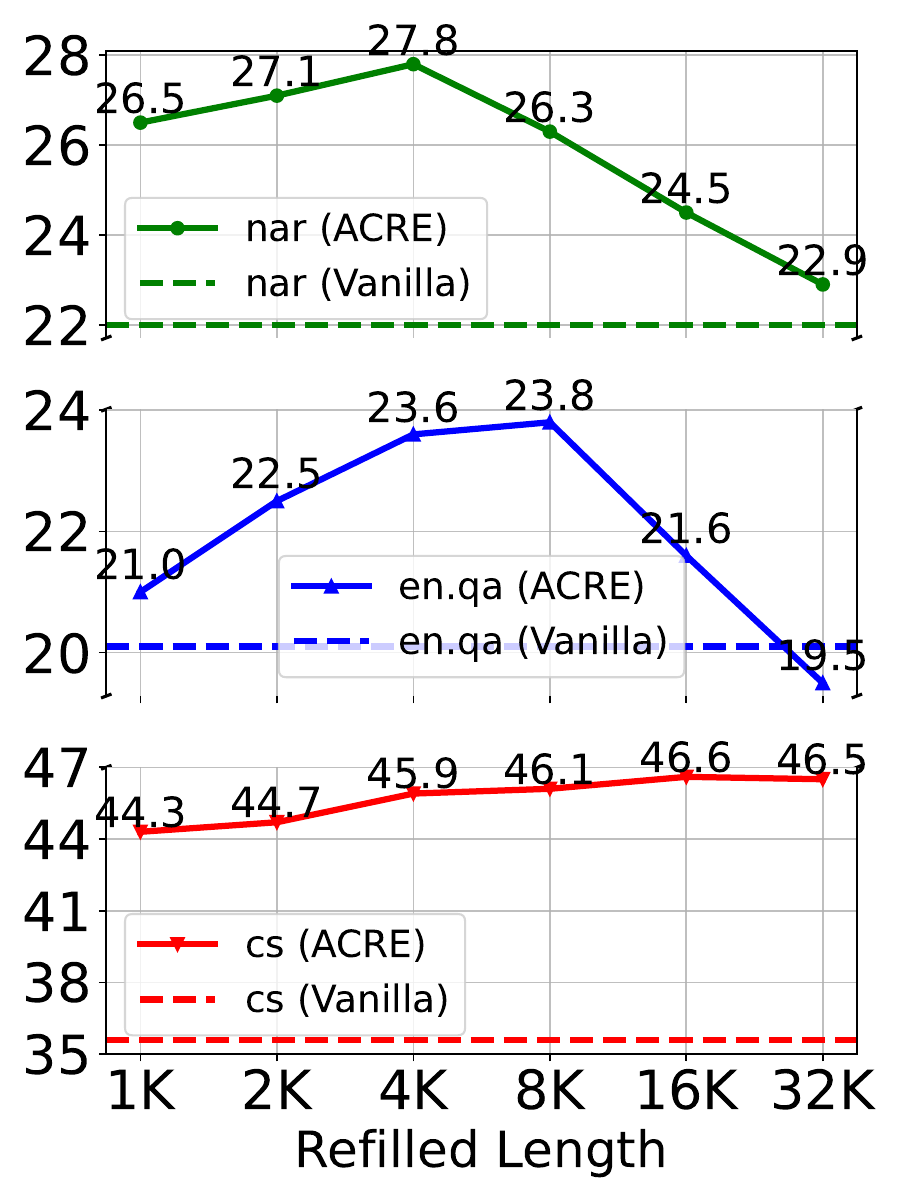}
    }
    \subfigure{
        \includegraphics[width=0.225\textwidth]{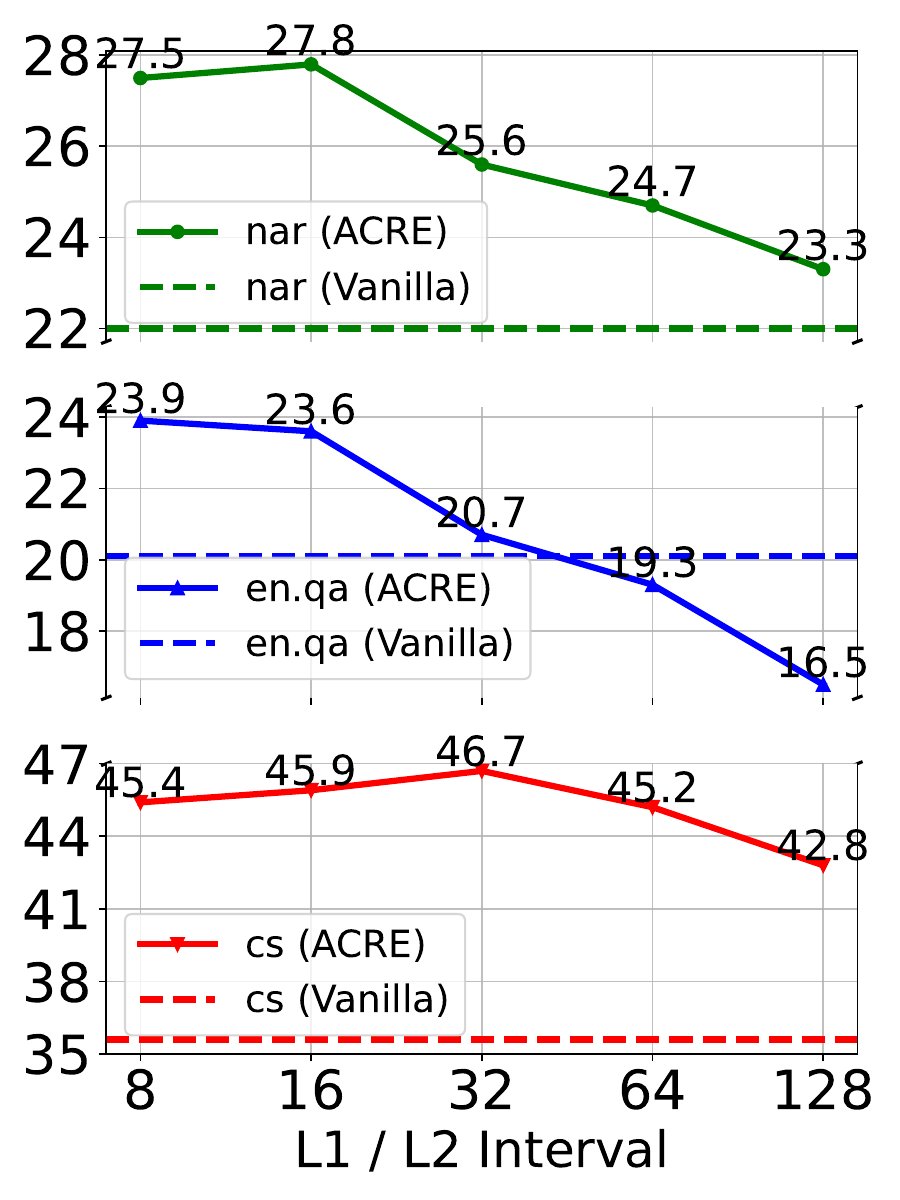}
    }

    \caption{Analysis of the maximum refilling length $\eta$~(left) and the impact of the L1/L2 interval $l$~(right).}
    \label{fig:ratio}

\end{figure}

(1) \textbf{Method Design and Model Selection}: 
Figure~\ref{fig:abl} presents ablation results across different LLMs and variations in model design.
First, we evaluate the role of training stages in model performance. Without the two-stage training process, ACRE reverts to a vanilla LLM, which performs significantly worse than ACRE. Stage-1 training enables ACRE to construct the bi-layer KV activation cache, thereby improving its long-context processing capabilities. When both stages are applied, ACRE achieves the best performance, demonstrating the effectiveness of its optimization design.

Second, to determine if ACRE’s effectiveness stems from its training data, we fine-tune a vanilla model using ACRE’s training data via SFT, producing \textit{SFT Vanilla}. While SFT improves the vanilla model by enhancing its QA capabilities, it still underperforms compared to ACRE. This highlights the unique advantages of ACRE’sdesign.

Lastly, we replace ACRE’s underlying LLM with Qwen2.5-7B (a scaled-up version of the same model) and Llama3.2-3B (a model of similar scale but different architecture). As shown in Figure~\ref{fig:abl}, ACRE’s design consistently proves effective across models of varying scales and architectures, confirming its generalizability.

(2) \textbf{Impact of Parameter Choice}:
As described in Section~\ref{sec:method}, ACRE’s performance may be influenced by two hyperparameters: the maximum refilling length of KV activations $\eta$ and the L1/L2 interval $l$. To investigate their impact, we conduct experiments with different values of $\eta$ and $l$. Figure~\ref{fig:ratio} presents the results of this analysis.

Specifically, in the left figure, we observe that the impact of the refilled activation length varies by task. For tasks with queries requiring explicit information (e.g., \texttt{nar} and \texttt{en.qa}), answer decoding relies on precise local information. Here, ACRE's performance peaks at a reasonable refilled length but declines as excessive refilling introduces noise, which biases the decoding process. Conversely, for tasks with queries requiring the integration of global information, ACRE's performance consistently improves with longer refilled lengths. This is because the L1 cache already provides global information, and additional refilled activations enhance local context.

The right figure shows the impact of the L1/L2 interval. We find that ACRE’s performance generally decreases as the L1/L2 interval increases. Larger intervals require L1 tokens to summarize more semantics from subsequent L2 tokens, potentially overloading the L1 cache. However, larger intervals result in a compact L1 KV cache, offering efficiency. In practical applications, users can adjust parameters to balance efficiency and effectiveness based on available resources.

In summary, ACRE outperforms directly using vanilla LLMs in most parameter settings, requiring significantly fewer computational resources while achieving higher efficiency.

\begin{table*}
\small
\caption{Efficiency comparison of Vanilla LLM, MInference, and ACRE. Peak GPU memory (\texttt{mem}, GiB), time latency (\texttt{lat}, seconds/query), and answer readability (\texttt{rdbl}) are evaluated using 20 samples with contexts over 1024K, truncated to target lengths, and a max generation length of 100 tokens. Tests are conducted on a single NVIDIA A800 80G GPU. Average scores are reported, with the best in each block highlighted in bold.} 
\begin{tabular}{lc@{\hspace{7.8pt}}c@{\hspace{7.8pt}}c@{\hspace{7.8pt}}|c@{\hspace{7.8pt}}c@{\hspace{7.8pt}}c@{\hspace{7.8pt}}|c@{\hspace{7.8pt}}c@{\hspace{7.8pt}}c@{\hspace{7.8pt}}|c@{\hspace{7.8pt}}c@{\hspace{7.8pt}}c@{\hspace{7.8pt}}|c@{\hspace{7.8pt}}c@{\hspace{7.8pt}}c@{\hspace{7.8pt}}}
    \toprule
Length  & \multicolumn{3}{c}{\textsc{64K}} & \multicolumn{3}{c}{\textsc{128K}} & \multicolumn{3}{c}{\textsc{256K}}  & \multicolumn{3}{c}{\textsc{512K}} & \multicolumn{3}{c}{\textsc{1024K}}  \\
\midrule
& \texttt{mem} & \texttt{lat} & \texttt{rdbl}  &  \texttt{mem} & \texttt{lat} & \texttt{rdbl} &  \texttt{mem} & \texttt{lat} & \texttt{rdbl} &  \texttt{mem} & \texttt{lat} & \texttt{rdbl}  &   \texttt{mem} & \texttt{lat} & \texttt{rdbl} \\
\midrule
 \multicolumn{6}{l}{\textsc{Qwen2.5-3B-instruct-128K}} \\
\midrule
Vanilla & 18.5 & 12.1 &\greentick& 27.9 & 36.3 &\greentick& 49.1 & 103.2 & \redcross& OOM & - & \redcross& OOM & - & \redcross\\
MInfer. & \textbf{15.5} & 29.2 &\greentick& \textbf{22.0} & 33.6 &\greentick& 28.0 & 57.1 & \redcross& \textbf{39.1} & 58.9 & \redcross& 47.2 & 79.6 &\redcross\\
ACRE    & 20.8 & \textbf{8.4} &\greentick& 23.0 & \textbf{14.3} &\greentick& \textbf{27.6} & \textbf{28.1} &\greentick & 44.3 & \textbf{48.2} & \greentick& \textbf{46.8} & \textbf{53.6} &\greentick\\
\midrule
 \multicolumn{6}{l}{\textsc{Qwen2.5-7B-instruct-128K}}\\
\midrule
Vanilla  & 31.9 & 21.2 &\greentick& 46.1 & 45.3 &\greentick& 78.3 & 129.6 &\redcross & OOM & - & \redcross& OOM & - &\redcross \\
MInfer.  & \textbf{27.9} & 29.1 &\greentick& \textbf{34.3} & 35.6 &\greentick& 48.1 & 81.2 & \redcross& 74.2 & 132.7 & \redcross& OOM & - &\redcross\\
ACRE     & 31.3 & \textbf{10.5} &\greentick& 35.1 & \textbf{18.0} &\greentick& \textbf{43.0} & \textbf{37.1} & \greentick& \textbf{72.1} & \textbf{85.6} & \greentick& \textbf{75.6} & \textbf{90.4} &\greentick\\

\bottomrule
\label{tab:effic}

\end{tabular}
\end{table*}

\subsection{Efficiency Analysis}
To evaluate ACRE's efficiency compared to baselines in processing long contexts at different scales, we conduct comparative experiments using the vanilla LLM, the efficient attention method MInference, and ACRE.

The results, presented in Table~\ref{tab:effic}, lead to the following conclusions:  
(1)~ACRE consistently processes long contexts at different scale with comparable or lower GPU resource usage. This efficiency is attributed to the bi-layer KV activation design, which avoids directly processing the full KV activations. 
(2)~ACRE's efficiency advantage becomes more pronounced with extremely long contexts (e.g., over 512K), where the vanilla LLM runs out of memory, and MInference faces a high risk of out of memory while require longer latency than ACRE. 
(3)~Thanks to its query-guided activation refilling mechanism, ACRE utilizes only the compact L1 KV activations and query-relevant L2 KV activations for answer decoding. This enables ACRE to process contexts longer than the native window of the LLM while maintaining answer quality. In contrast, baseline models generate nonsensical answers when exceeding LLM's native context length.

In summary, ACRE demonstrates significant advantages in handling long contexts efficiently and reliably compared to baseline methods.

\section{Related Work}
Long-context processing is a critical capability of LLMs~\cite{zhao2024surveylargelanguagemodels}. The most fundamental approach to enhancing this ability is training LLMs on long texts, either sampled from raw corpora or synthesized~\cite{xiong2023effective,mohtashami2024random,fu2024data}. Consequently, the native context window of popular LLMs has increased significantly, from the earlier 4K to the current 128K~\cite{peng2023yarn, llama, qwen2}. 

In addition to directly increasing the context window, some methods employ strategic positional encoding to enable LLMs to process contexts longer than their native window, as demonstrated by~\cite{chen2023extending, song2023hierarchical, liu2023scaling, jin2024llmmaybelonglmselfextend}. However, when processing long contexts, LLMs generate large key-value (KV) activations, which consume substantial resources and reduce efficiency. To address this, many works aim to make KV activations more compact and lightweight~\cite{liukivi,xu2024think}. 
For example, KIVI focuses on reducing the precision of KV activations to 2-bit, resulting in significantly lighter KV representations~\cite{liukivi}. 
Other methods selectively attend to a small portion of KV activations through compression or sparse attention mechanisms. For instance, StreamingLLM proposes attending only to recent tokens and sink tokens to maintain compact KV activations~\cite{xiao2024efficientstreaminglanguagemodels}, similar idea also adopted by~\cite{li2024snapkv,yao2024theoretical, jiang2024minference, zhang2024soaring}. 
Beyond optimizing KV activations, alternative methods such as agent-based approaches~\cite{qian2024longllmsnecessitylongcontexttasks,lee2024humaninspiredreadingagentgist} and retrieval-augmented generation~\cite{longctx,zhu2024largelanguagemodelsinformation,zhou2024trustworthiness} have been applied to facilitate long-context processing. These methods split the long context into chunks and retrieve evidence using retrievers or agents. They work well for explicit queries but struggle with implicit ones requiring full-context aggregation.

Most existing methods  either compact global KV activations into a lightweight form or prune them into shorter forms, often failing to balance global perspective with local informativeness. This limitation can compromise performance in information-seeking scenarios, where information needs may dynamically range from global to local.
\section{Conclusion}
In this paper, we propose a method, ACRE, designed to adapt to the dynamic information needs of long-context information-seeking tasks. ACRE constructs a bi-layer KV activation cache structure for long contexts, where the L1 KV cache stores compact, global information, and the L2 KV cache captures detailed, local information. Using query-guided activation refilling, ACRE identifies query-specific evidence from the L2 KV cache and refills this local information into the L1 KV cache, resulting in nested KV activations that effectively combine a global perspective with local details. Through experiments on a wide range of information-seeking datasets, we demonstrate the effectiveness of ACRE in simultaneously improving the performance and efficiency of long-context processing for information-seeking tasks.

\section*{Limitation}
In this paper, we propose ACRE, a method designed to adapt to the dynamic information needs of long-context information-seeking tasks. ACRE constructs a bi-layer KV activation cache to balance global context perception and local detail preservation, leveraging query-guided activation refilling to enhance performance and efficiency. While ACRE demonstrates significant advancements, several limitations are worth noting:

(1)~Our method is primarily designed for information-seeking tasks, a major subset of long-context processing. This focus is largely driven by the availability of training data, as information-seeking tasks benefit from abundant QA datasets. While ACRE has the potential to adapt to general long-context tasks, further exploration with diverse task-specific data would be necessary to validate its broader applicability.

(2)~ACRE introduces additional parameters for constructing the bi-layer KV cache, increasing the model size. For example, using Qwen2.5-3B-Instruct, ACRE adds approximately 17.2\% more parameters, requiring additional GPU memory to load the model. However, in long-context tasks, the majority of GPU memory is consumed by KV activations rather than model parameters. Our efficiency analysis confirms that ACRE reduces overall GPU memory consumption when processing long contexts, mitigating this limitation to some extent.

(3)~A portion of our training data is synthetically generated by commercial LLMs (e.g. GPT-4), which may introduce biases inherited from the original corpus or the LLMs used. While such biases could impact performance, many current commercial LLMs incorporate robust safeguards that help mitigate these issues. Nonetheless, addressing potential biases in synthetic data remains an area for future improvement.

\bibliography{custom}

\begin{thebibliography}{43}
\providecommand{\natexlab}[1]{#1}

\bibitem[{Bai et~al.(2024)Bai, Lv, Zhang, Lyu, Tang, Huang, Du, Liu, Zeng, Hou, Dong, Tang, and Li}]{bai2023longbench}
Yushi Bai, Xin Lv, Jiajie Zhang, Hongchang Lyu, Jiankai Tang, Zhidian Huang, Zhengxiao Du, Xiao Liu, Aohan Zeng, Lei Hou, Yuxiao Dong, Jie Tang, and Juanzi Li. 2024.
\newblock \href {https://doi.org/10.18653/V1/2024.ACL-LONG.172} {Longbench: {A} bilingual, multitask benchmark for long context understanding}.
\newblock In \emph{Proceedings of the 62nd Annual Meeting of the Association for Computational Linguistics (Volume 1: Long Papers), {ACL} 2024, Bangkok, Thailand, August 11-16, 2024}, pages 3119--3137. Association for Computational Linguistics.

\bibitem[{Chan et~al.(2024)Chan, Xu, Yuan, Luo, Xue, Guo, and Fu}]{chan2024rqraglearningrefinequeries}
Chi{-}Min Chan, Chunpu Xu, Ruibin Yuan, Hongyin Luo, Wei Xue, Yike Guo, and Jie Fu. 2024.
\newblock \href {https://doi.org/10.48550/ARXIV.2404.00610} {{RQ-RAG:} learning to refine queries for retrieval augmented generation}.
\newblock \emph{CoRR}, abs/2404.00610.

\bibitem[{Chen et~al.(2023{\natexlab{a}})Chen, Xiao, Zhang, Luo, Lian, and Liu}]{bge_m3}
Jianlv Chen, Shitao Xiao, Peitian Zhang, Kun Luo, Defu Lian, and Zheng Liu. 2023{\natexlab{a}}.
\newblock \href {https://arxiv.org/abs/2309.07597} {Bge m3-embedding: Multi-lingual, multi-functionality, multi-granularity text embeddings through self-knowledge distillation}.
\newblock \emph{Preprint}, arXiv:2309.07597.

\bibitem[{Chen et~al.(2023{\natexlab{b}})Chen, Wong, Chen, and Tian}]{chen2023extending}
Shouyuan Chen, Sherman Wong, Liangjian Chen, and Yuandong Tian. 2023{\natexlab{b}}.
\newblock Extending context window of large language models via positional interpolation.
\newblock \emph{arXiv preprint arXiv:2306.15595}.

\bibitem[{Chen et~al.(2024)Chen, Qian, Tang, Lai, Liu, Han, and Jia}]{chen2024longloraefficientfinetuninglongcontext}
Yukang Chen, Shengju Qian, Haotian Tang, Xin Lai, Zhijian Liu, Song Han, and Jiaya Jia. 2024.
\newblock \href {https://arxiv.org/abs/2309.12307} {Longlora: Efficient fine-tuning of long-context large language models}.
\newblock \emph{Preprint}, arXiv:2309.12307.

\bibitem[{Dasigi et~al.(2021)Dasigi, Lo, Beltagy, Cohan, Smith, and Gardner}]{dasigi2021dataset}
Pradeep Dasigi, Kyle Lo, Iz~Beltagy, Arman Cohan, Noah~A Smith, and Matt Gardner. 2021.
\newblock A dataset of information-seeking questions and answers anchored in research papers.
\newblock In \emph{Proceedings of the 2021 Conference of the North American Chapter of the Association for Computational Linguistics: Human Language Technologies}, pages 4599--4610.

\bibitem[{Dong et~al.(2023)Dong, Tang, Li, and Zhao}]{dong2023survey}
Zican Dong, Tianyi Tang, Lunyi Li, and Wayne~Xin Zhao. 2023.
\newblock A survey on long text modeling with transformers.
\newblock \emph{arXiv preprint arXiv:2302.14502}.

\bibitem[{Fu et~al.(2024)Fu, Panda, Niu, Yue, Hajishirzi, Kim, and Peng}]{fu2024data}
Yao Fu, Rameswar Panda, Xinyao Niu, Xiang Yue, Hannaneh Hajishirzi, Yoon Kim, and Hao Peng. 2024.
\newblock \href {https://arxiv.org/abs/2402.10171} {Data engineering for scaling language models to 128k context}.
\newblock \emph{Preprint}, arXiv:2402.10171.

\bibitem[{Gao et~al.(2024)Gao, Xiong, Gao, Jia, Pan, Bi, Dai, Sun, Guo, Wang, and Wang}]{gao2024retrievalaugmented}
Yunfan Gao, Yun Xiong, Xinyu Gao, Kangxiang Jia, Jinliu Pan, Yuxi Bi, Yi~Dai, Jiawei Sun, Qianyu Guo, Meng Wang, and Haofen Wang. 2024.
\newblock \href {https://arxiv.org/abs/2312.10997} {Retrieval-augmented generation for large language models: A survey}.
\newblock \emph{Preprint}, arXiv:2312.10997.

\bibitem[{Ho et~al.(2020)Ho, Duong~Nguyen, Sugawara, and Aizawa}]{ho-etal-2020-constructing}
Xanh Ho, Anh-Khoa Duong~Nguyen, Saku Sugawara, and Akiko Aizawa. 2020.
\newblock \href {https://doi.org/10.18653/v1/2020.coling-main.580} {Constructing a multi-hop {QA} dataset for comprehensive evaluation of reasoning steps}.
\newblock In \emph{Proceedings of the 28th International Conference on Computational Linguistics}, pages 6609--6625, Barcelona, Spain (Online). International Committee on Computational Linguistics.

\bibitem[{Jiang et~al.(2024)Jiang, Li, Zhang, Wu, Luo, Ahn, Han, Abdi, Li, Lin, Yang, and Qiu}]{jiang2024minference}
Huiqiang Jiang, Yucheng Li, Chengruidong Zhang, Qianhui Wu, Xufang Luo, Surin Ahn, Zhenhua Han, Amir~H Abdi, Dongsheng Li, Chin-Yew Lin, Yuqing Yang, and Lili Qiu. 2024.
\newblock Minference 1.0: Accelerating pre-filling for long-context llms via dynamic sparse attention.
\newblock \emph{arXiv preprint arXiv:2407.02490}.

\bibitem[{Jin et~al.(2024)Jin, Han, Yang, Jiang, Liu, Chang, Chen, and Hu}]{jin2024llmmaybelonglmselfextend}
Hongye Jin, Xiaotian Han, Jingfeng Yang, Zhimeng Jiang, Zirui Liu, Chia-Yuan Chang, Huiyuan Chen, and Xia Hu. 2024.
\newblock \href {https://arxiv.org/abs/2401.01325} {Llm maybe longlm: Self-extend llm context window without tuning}.
\newblock \emph{Preprint}, arXiv:2401.01325.

\bibitem[{Kocisk{\'{y}} et~al.(2018)Kocisk{\'{y}}, Schwarz, Blunsom, Dyer, Hermann, Melis, and Grefenstette}]{kočiský2017narrativeqa}
Tom{\'{a}}s Kocisk{\'{y}}, Jonathan Schwarz, Phil Blunsom, Chris Dyer, Karl~Moritz Hermann, G{\'{a}}bor Melis, and Edward Grefenstette. 2018.
\newblock \href {https://doi.org/10.1162/TACL\_A\_00023} {The narrativeqa reading comprehension challenge}.
\newblock \emph{Trans. Assoc. Comput. Linguistics}, 6:317--328.

\bibitem[{Lee et~al.(2024)Lee, Chen, Furuta, Canny, and Fischer}]{lee2024humaninspiredreadingagentgist}
Kuang{-}Huei Lee, Xinyun Chen, Hiroki Furuta, John~F. Canny, and Ian Fischer. 2024.
\newblock \href {https://openreview.net/forum?id=OTmcsyEO5G} {A human-inspired reading agent with gist memory of very long contexts}.
\newblock In \emph{Forty-first International Conference on Machine Learning, {ICML} 2024, Vienna, Austria, July 21-27, 2024}. OpenReview.net.

\bibitem[{Li et~al.(2024)Li, Huang, Yang, Venkitesh, Locatelli, Ye, Cai, Lewis, and Chen}]{li2024snapkv}
Yuhong Li, Yingbing Huang, Bowen Yang, Bharat Venkitesh, Acyr Locatelli, Hanchen Ye, Tianle Cai, Patrick Lewis, and Deming Chen. 2024.
\newblock Snapkv: Llm knows what you are looking for before generation.
\newblock \emph{arXiv preprint arXiv:2404.14469}.

\bibitem[{Liu et~al.(2023)Liu, Yan, An, Qiu, and Lin}]{liu2023scaling}
Xiaoran Liu, Hang Yan, Chenxin An, Xipeng Qiu, and Dahua Lin. 2023.
\newblock Scaling laws of rope-based extrapolation.
\newblock In \emph{The Twelfth International Conference on Learning Representations}.

\bibitem[{Liu et~al.(2024)Liu, Yuan, Jin, Zhong, Xu, Braverman, Chen, and Hu}]{liukivi}
Zirui Liu, Jiayi Yuan, Hongye Jin, Shaochen Zhong, Zhaozhuo Xu, Vladimir Braverman, Beidi Chen, and Xia Hu. 2024.
\newblock Kivi: A tuning-free asymmetric 2bit quantization for kv cache.
\newblock In \emph{Forty-first International Conference on Machine Learning}.

\bibitem[{MetaAI(2024)}]{grattafiori2024llama3herdmodels}
MetaAI. 2024.
\newblock \href {https://arxiv.org/abs/2407.21783} {The llama 3 herd of models}.
\newblock \emph{Preprint}, arXiv:2407.21783.

\bibitem[{Mohtashami and Jaggi(2024)}]{mohtashami2024random}
Amirkeivan Mohtashami and Martin Jaggi. 2024.
\newblock Random-access infinite context length for transformers.
\newblock \emph{Advances in Neural Information Processing Systems}, 36.

\bibitem[{OpenAI(2023)}]{gpt-4}
OpenAI. 2023.
\newblock Gpt-4 technical report.
\newblock \url{https://cdn.openai.com/papers/gpt-4.pdf}.

\bibitem[{Peng et~al.(2023)Peng, Quesnelle, Fan, and Shippole}]{peng2023yarn}
Bowen Peng, Jeffrey Quesnelle, Honglu Fan, and Enrico Shippole. 2023.
\newblock Yarn: Efficient context window extension of large language models.
\newblock In \emph{The Twelfth International Conference on Learning Representations}.

\bibitem[{Qian et~al.(2025{\natexlab{a}})Qian, Liu, Gao, Wang, Lian, and Dou}]{qian2025hawkbench}
Hongjin Qian, Zheng Liu, Chao Gao, Yankai Wang, Defu Lian, and Zhicheng Dou. 2025{\natexlab{a}}.
\newblock Hawkbench: Investigating resilience of rag methods on stratified information-seeking tasks.
\newblock \emph{arXiv preprint arXiv:2502.13465}.

\bibitem[{Qian et~al.(2025{\natexlab{b}})Qian, Liu, Zhang, Mao, Lian, Dou, and Huang}]{qian2024memoragmovingnextgenrag}
Hongjin Qian, Zheng Liu, Peitian Zhang, Kelong Mao, Defu Lian, Zhicheng Dou, and Tiejun Huang. 2025{\natexlab{b}}.
\newblock Memorag: Boosting long context processing with global memory-enhanced retrieval augmentation.
\newblock In \emph{Proceedings of the ACM on Web Conference 2025}, pages 2366--2377.

\bibitem[{Qian et~al.(2024)Qian, Liu, Zhang, Mao, Zhou, Chen, and Dou}]{qian2024longllmsnecessitylongcontexttasks}
Hongjin Qian, Zheng Liu, Peitian Zhang, Kelong Mao, Yujia Zhou, Xu~Chen, and Zhicheng Dou. 2024.
\newblock \href {https://arxiv.org/abs/2405.15318} {Are long-llms a necessity for long-context tasks?}
\newblock \emph{Preprint}, arXiv:2405.15318.

\bibitem[{Soboleva et~al.(2023)Soboleva, Al-Khateeb, Myers, Steeves, Hestness, and Dey}]{together2023redpajama}
Daria Soboleva, Faisal Al-Khateeb, Robert Myers, Jacob~R Steeves, Joel Hestness, and Nolan Dey. 2023.
\newblock \href {https://huggingface.co/datasets/cerebras/SlimPajama-627B} {{SlimPajama: A 627B token cleaned and deduplicated version of RedPajama}}.

\bibitem[{Song et~al.(2023)Song, Oh, Mo, Kim, Yun, Ha, and Shin}]{song2023hierarchical}
Woomin Song, Seunghyuk Oh, Sangwoo Mo, Jaehyung Kim, Sukmin Yun, Jung-Woo Ha, and Jinwoo Shin. 2023.
\newblock Hierarchical context merging: Better long context understanding for pre-trained llms.
\newblock In \emph{The Twelfth International Conference on Learning Representations}.

\bibitem[{Touvron et~al.(2023)Touvron, Martin, Stone, Albert, Almahairi, Babaei, Bashlykov, Batra, Bhargava, Bhosale et~al.}]{llama}
Hugo Touvron, Louis Martin, Kevin Stone, Peter Albert, Amjad Almahairi, Yasmine Babaei, Nikolay Bashlykov, Soumya Batra, Prajjwal Bhargava, Shruti Bhosale, et~al. 2023.
\newblock Llama 2: Open foundation and fine-tuned chat models.
\newblock \emph{arXiv preprint arXiv:2307.09288}.

\bibitem[{Trivedi et~al.(2022)Trivedi, Balasubramanian, Khot, and Sabharwal}]{trivedi2022musique}
Harsh Trivedi, Niranjan Balasubramanian, Tushar Khot, and Ashish Sabharwal. 2022.
\newblock Musique: Multihop questions via single-hop question composition.
\newblock \emph{Transactions of the Association for Computational Linguistics}, 10:539--554.

\bibitem[{Vaswani et~al.(2023)Vaswani, Shazeer, Parmar, Uszkoreit, Jones, Gomez, Kaiser, and Polosukhin}]{vaswani2023attentionneed}
Ashish Vaswani, Noam Shazeer, Niki Parmar, Jakob Uszkoreit, Llion Jones, Aidan~N. Gomez, Lukasz Kaiser, and Illia Polosukhin. 2023.
\newblock \href {https://arxiv.org/abs/1706.03762} {Attention is all you need}.
\newblock \emph{Preprint}, arXiv:1706.03762.

\bibitem[{Xiao et~al.(2024)Xiao, Tian, Chen, Han, and Lewis}]{xiao2024efficientstreaminglanguagemodels}
Guangxuan Xiao, Yuandong Tian, Beidi Chen, Song Han, and Mike Lewis. 2024.
\newblock \href {https://arxiv.org/abs/2309.17453} {Efficient streaming language models with attention sinks}.
\newblock \emph{Preprint}, arXiv:2309.17453.

\bibitem[{Xiong et~al.(2024)Xiong, Liu, Molybog, Zhang, Bhargava, Hou, Martin, Rungta, Sankararaman, Oguz, Khabsa, Fang, Mehdad, Narang, Malik, Fan, Bhosale, Edunov, Lewis, Wang, and Ma}]{xiong2023effective}
Wenhan Xiong, Jingyu Liu, Igor Molybog, Hejia Zhang, Prajjwal Bhargava, Rui Hou, Louis Martin, Rashi Rungta, Karthik~Abinav Sankararaman, Barlas Oguz, Madian Khabsa, Han Fang, Yashar Mehdad, Sharan Narang, Kshitiz Malik, Angela Fan, Shruti Bhosale, Sergey Edunov, Mike Lewis, Sinong Wang, and Hao Ma. 2024.
\newblock \href {https://doi.org/10.18653/V1/2024.NAACL-LONG.260} {Effective long-context scaling of foundation models}.
\newblock In \emph{Proceedings of the 2024 Conference of the North American Chapter of the Association for Computational Linguistics: Human Language Technologies (Volume 1: Long Papers), {NAACL} 2024, Mexico City, Mexico, June 16-21, 2024}, pages 4643--4663. Association for Computational Linguistics.

\bibitem[{Xu et~al.(2023)Xu, Ping, Wu, McAfee, Zhu, Liu, Subramanian, Bakhturina, Shoeybi, and Catanzaro}]{longctx}
Peng Xu, Wei Ping, Xianchao Wu, Lawrence McAfee, Chen Zhu, Zihan Liu, Sandeep Subramanian, Evelina Bakhturina, Mohammad Shoeybi, and Bryan Catanzaro. 2023.
\newblock \href {https://doi.org/10.48550/arxiv.2310.03025} {{Retrieval meets Long Context Large Language Models}}.
\newblock \emph{arXiv}.
\newblock Experimental.

\bibitem[{Xu et~al.(2024)Xu, Jie, Dong, Wang, Lu, Zhou, Saha, Xiong, and Sahoo}]{xu2024think}
Yuhui Xu, Zhanming Jie, Hanze Dong, Lei Wang, Xudong Lu, Aojun Zhou, Amrita Saha, Caiming Xiong, and Doyen Sahoo. 2024.
\newblock Think: Thinner key cache by query-driven pruning.
\newblock \emph{arXiv preprint arXiv:2407.21018}.

\bibitem[{Yang et~al.(2024)Yang, Yang, Hui, Zheng, Yu, Zhou, Li, Li, Liu, Huang et~al.}]{qwen2}
An~Yang, Baosong Yang, Binyuan Hui, Bo~Zheng, Bowen Yu, Chang Zhou, Chengpeng Li, Chengyuan Li, Dayiheng Liu, Fei Huang, et~al. 2024.
\newblock Qwen2 technical report.
\newblock \emph{arXiv preprint arXiv:2407.10671}.

\bibitem[{Yao et~al.(2024)Yao, Qian, Hu, Xu, Liu, Liu, Luan, and Wang}]{yao2024theoretical}
Xinhao Yao, Hongjin Qian, Xiaolin Hu, Gengze Xu, Yong Liu, Wei Liu, Jian Luan, and Bin Wang. 2024.
\newblock Theoretical insights into fine-tuning attention mechanism: Generalization and optimization.
\newblock \emph{arXiv preprint arXiv:2410.02247}.

\bibitem[{Yoon et~al.(2024)Yoon, Lee, Hwang, Jeong, and Kang}]{yoon2024compactcompressingretrieveddocuments}
Chanwoong Yoon, Taewhoo Lee, Hyeon Hwang, Minbyul Jeong, and Jaewoo Kang. 2024.
\newblock \href {https://arxiv.org/abs/2407.09014} {Compact: Compressing retrieved documents actively for question answering}.
\newblock \emph{Preprint}, arXiv:2407.09014.

\bibitem[{Zhang et~al.(2024{\natexlab{a}})Zhang, Liu, Xiao, Shao, Ye, and Dou}]{zhang2024soaring}
Peitian Zhang, Zheng Liu, Shitao Xiao, Ninglu Shao, Qiwei Ye, and Zhicheng Dou. 2024{\natexlab{a}}.
\newblock Soaring from 4k to 400k: Extending llm's context with activation beacon.
\newblock \emph{arXiv preprint arXiv:2401.03462}.

\bibitem[{Zhang et~al.(2024{\natexlab{b}})Zhang, Shao, Liu, Xiao, Qian, Ye, and Dou}]{zhang2024extending}
Peitian Zhang, Ninglu Shao, Zheng Liu, Shitao Xiao, Hongjin Qian, Qiwei Ye, and Zhicheng Dou. 2024{\natexlab{b}}.
\newblock \href {https://arxiv.org/abs/2404.19553} {Extending llama-3's context ten-fold overnight}.
\newblock \emph{Preprint}, arXiv:2404.19553.

\bibitem[{Zhang et~al.(2024{\natexlab{c}})Zhang, Chen, Hu, Xu, Chen, Hao, Han, Thai, Wang, Liu, and Sun}]{zhang2024inftybench}
Xinrong Zhang, Yingfa Chen, Shengding Hu, Zihang Xu, Junhao Chen, Moo~Khai Hao, Xu~Han, Zhen~Leng Thai, Shuo Wang, Zhiyuan Liu, and Maosong Sun. 2024{\natexlab{c}}.
\newblock \href {https://doi.org/10.18653/V1/2024.ACL-LONG.814} {{\i}nftybench: Extending long context evaluation beyond 100k tokens}.
\newblock In \emph{Proceedings of the 62nd Annual Meeting of the Association for Computational Linguistics (Volume 1: Long Papers), {ACL} 2024, Bangkok, Thailand, August 11-16, 2024}, pages 15262--15277. Association for Computational Linguistics.

\bibitem[{Zhang et~al.(2023)Zhang, Sheng, Zhou, Chen, Zheng, Cai, Song, Tian, Ré, Barrett, Wang, and Chen}]{zhang2023h2oheavyhitteroracleefficient}
Zhenyu Zhang, Ying Sheng, Tianyi Zhou, Tianlong Chen, Lianmin Zheng, Ruisi Cai, Zhao Song, Yuandong Tian, Christopher Ré, Clark Barrett, Zhangyang Wang, and Beidi Chen. 2023.
\newblock \href {https://arxiv.org/abs/2306.14048} {H$_2$o: Heavy-hitter oracle for efficient generative inference of large language models}.
\newblock \emph{Preprint}, arXiv:2306.14048.

\bibitem[{Zhao et~al.(2024)Zhao, Zhou, Li, Tang, Wang, Hou, Min, Zhang, Zhang, Dong, Du, Yang, Chen, Chen, Jiang, Ren, Li, Tang, Liu, Liu, Nie, and Wen}]{zhao2024surveylargelanguagemodels}
Wayne~Xin Zhao, Kun Zhou, Junyi Li, Tianyi Tang, Xiaolei Wang, Yupeng Hou, Yingqian Min, Beichen Zhang, Junjie Zhang, Zican Dong, Yifan Du, Chen Yang, Yushuo Chen, Zhipeng Chen, Jinhao Jiang, Ruiyang Ren, Yifan Li, Xinyu Tang, Zikang Liu, Peiyu Liu, Jian-Yun Nie, and Ji-Rong Wen. 2024.
\newblock \href {https://arxiv.org/abs/2303.18223} {A survey of large language models}.
\newblock \emph{Preprint}, arXiv:2303.18223.

\bibitem[{Zhou et~al.(2024)Zhou, Liu, Li, Jin, Qian, Liu, Li, Dou, Ho, and Yu}]{zhou2024trustworthiness}
Yujia Zhou, Yan Liu, Xiaoxi Li, Jiajie Jin, Hongjin Qian, Zheng Liu, Chaozhuo Li, Zhicheng Dou, Tsung-Yi Ho, and Philip~S Yu. 2024.
\newblock Trustworthiness in retrieval-augmented generation systems: A survey.
\newblock \emph{arXiv preprint arXiv:2409.10102}.

\bibitem[{Zhu et~al.(2024)Zhu, Yuan, Wang, Liu, Liu, Deng, Chen, Dou, and Wen}]{zhu2024largelanguagemodelsinformation}
Yutao Zhu, Huaying Yuan, Shuting Wang, Jiongnan Liu, Wenhan Liu, Chenlong Deng, Haonan Chen, Zhicheng Dou, and Ji-Rong Wen. 2024.
\newblock \href {https://arxiv.org/abs/2308.07107} {Large language models for information retrieval: A survey}.
\newblock \emph{Preprint}, arXiv:2308.07107.

\end{thebibliography}

\appendix

\section{Implementation details}
\label{sec:imp}
For ACRE training, in stage 1, we sample long text spans from the RedPajama~\cite{together2023redpajama} dataset to create a training set of 2 billion tokens. The sampled text lengths are limited to a minimum of 4K and a maximum of 64K tokens. We randomly choose L1/L2 interval from $l \in \{8, 16, 32, 64, 128\}$. The model is trained for one epoch with a batch size of 8 and a learning rate of $5 \times 10^{-5}$. 
In stage 2, we collect 28,400 QA SFT data points from LongAlpaca~\cite{chen2024longloraefficientfinetuninglongcontext} and synthetic data from~\cite{zhang2024soaring,qian2024memoragmovingnextgenrag}. We apply the same L1 token insertion strategy during training. The model is trained for three epochs with a batch size of 8 and a learning rate of $1 \times 10^{-5}$ for two epochs. Stage-1 training takes around 7 hours while stage-2 training takes around 13 hours. 

During the two-stage training process, we optimize only the newly initialized parameters, keeping the original parameters frozen. The number of trainable parameters varies depending on the model. For instance:
(1)~When using Qwen2.5-3B-instruct, ACRE has around 503M trainable parameters, accounting for 17.2\% of the original parameters.
(2)~When using Llama3.2-3B-instruct, ACRE has around 780M trainable parameters, accounting for 25.6\% of the original parameters.
This difference arises from variations in the implementation of multi-head attention.

\begin{cvbox}[Prompt for Bi-Layer KV Cache Construction]
You are provided with a long article. Read the article carefully. 

After reading, you will be asked to perform specific tasks based on the content of the article.

Now, the article begins:

**Article Content:** [context]

The article ends here.

Next, follow the instructions provided to complete the tasks.
\label{tab:prompt}
\end{cvbox}

For the main experiments, we configure ACRE with an L1/L2 interval $l$ of 16, a maximum refilling length $\eta$ of 4,096, and the maximum working context window $\gW$ of 32K tokens. For the Bi-Layer KV Cache construction, we utilize the following prompt. During the Query-Guided Activation Refilling process, we adopt task-specific prompts from the official benchmark repositories, without inserting the context into the task prompt.

For RAG, RQ-RAG, and MemoRAG, we employ BGE-M3~\cite{bge_m3} as the retriever and set the hit number to 5. For methods that divide the long context into chunks, we use the \href{https://pypi.org/project/semantic-text-splitter/}{semantic-text-splitter} tool, chunking the context to a maximum length of 512 tokens. 

For KIVI, we quantize the KV activations to 4-bit precision. For Beacon, we use the official training code to fine-tune Qwen2.5-3B-Instruct, setting the compression ratio to 8 during inference. For SelfExtend, we set the group size to 32 and the window size to 2048, which is approximate by the official recommended strategy. For StreamingLLM, we use the SinkCache implementation from Transformers, configuring the window size to 4096 and the number of sink tokens to 8. Lastly, for MemoRAG, we utilize the officially released \href{https://huggingface.co/TommyChien/memorag-qwen2-7b-inst}{memorag-qwen2-7b-inst} as the memory model.

All methods are evaluated using the task prompts provided in the official repositories of their corresponding benchmarks\footnote{LongBench: \url{https://github.com/THUDM/LongBench}, InfiniteBench: \url{https://github.com/OpenBMB/InfiniteBench}}. Additionally, we use the same generation hyper-parameters (task-dependent) for ACRE and all baseline models.

All training and evaluation experiments were conducted using 8 NVIDIA A800-80G GPUs.

\end{document}